\documentclass[sigconf,anonymous=false]{acmart}

\AtBeginDocument{%
  \providecommand\BibTeX{{%
    \normalfont B\kern-0.5em{\scshape i\kern-0.25em b}\kern-0.8em\TeX}}}

\setcopyright{acmcopyright}

\copyrightyear{2020}
\acmYear{2020}
\setcopyright{rightsretained}
\acmConference[GECCO '20 Companion]{Genetic and Evolutionary Computation Conference Companion}{July 8--12, 2020}{Canc\'{u}n, Mexico}
\acmBooktitle{Genetic and Evolutionary Computation Conference Companion (GECCO '20 Companion), July 8--12, 2020, Canc\'{u}n, Mexico}\acmDOI{10.1145/3377929.3389974}
\acmISBN{978-1-4503-7127-8/20/07}

\usepackage{siunitx}

\usepackage[colorinlistoftodos,disable]{todonotes}
\usepackage{censor}
\def\censorcolor{white}

\let\svcensorrule\censorrule
\renewcommand\censorrule[1]{%
\textcolor{\censorcolor}{\svcensorrule{#1}}}

\begin{document}

\title[Towards Realistic Optimization Benchmarks: A Questionnaire on the Properties of Real-World Problems]{Towards Realistic Optimization Benchmarks:\\ A Questionnaire on the Properties of Real-World Problems}

\author[K. van der Blom]{Koen van der Blom}
\email{k.van.der.blom@liacs.leidenuniv.nl}
\orcid{0000-0002-4653-0707}
\affiliation{%
  \institution{
  Leiden University}
  \streetaddress{Niels Bohrweg 1}
  \city{Leiden}
  \country{The Netherlands}
  \postcode{2333 CA}
}
  
\author[T. M. Deist]{Timo M. Deist}
\email{timo.deist@cwi.nl}
\orcid{}
\affiliation{%
  \institution{Centrum Wiskunde \& Informatica}
  \city{Amsterdam}
  \country{The Netherlands}
}

\author[T. Tu\v{s}ar]{Tea Tu\v{s}ar}
\email{tea.tusar@ijs.si}
\orcid{}
\affiliation{%
  \institution{Jozef Stefan Institute}
  \city{Ljubljana}
  \country{Slovenia}
}

\author[M. Marchi]{Mariapia Marchi}
\email{marchi@esteco.com}
\orcid{}
\affiliation{%
  \institution{ESTECO SpA}
  \city{Trieste}
  \country{Italy}
}

\author[Y. Nojima]{Yusuke Nojima}
\email{nojima@cs.osakafu-u.ac.jp}
\orcid{}
\affiliation{%
  \institution{Osaka Prefecture University}
  \city{Sakai, Osaka}
  \country{Japan}
}

\author[A. Oyama]{Akira Oyama}
\email{oyama@flab.isas.jaxa.jp}
\orcid{}
\affiliation{%
  \institution{Japan Aerospace Exploration Agency}
  \city{Sagamihara}
  \country{Japan}
}

\author[V. Volz]{Vanessa Volz}
\email{vanessa@modl.ai}
\affiliation{%
  \institution{modl.ai}
  \streetaddress{Nørrebrogade 184, 1}
  \city{Copenhagen, Denmark} 
}

\author[B. Naujoks]{Boris Naujoks}
\email{boris.naujoks@th-koeln.de}
\orcid{}
\affiliation{%
  \institution{TH K\"{o}ln}
  \city{Gummersbach}
  \country{Germany}
}


\begin{abstract}
Benchmarks are a useful tool for empirical performance comparisons. 
However, one of the main shortcomings of existing benchmarks is that it remains largely unclear how they relate to real-world problems. What does an algorithm's performance on a benchmark say about its potential on a specific real-world problem?
This work aims to identify properties of real-world problems through a questionnaire on real-world single-, multi-, and many-objective optimization problems.
Based on initial responses, a few challenges that have to be considered in the design of realistic benchmarks can already be identified.
A key point for future work is to gather more responses to the questionnaire to allow an analysis of common combinations of properties. In turn, such common combinations can then be included in improved benchmark suites. To gather more data, the reader is invited to participate in the questionnaire at: \url{https://tinyurl.com/opt-survey}
\end{abstract}



\maketitle

\todo[inline]{Update doi+isbn after completing copyright process}

\todo[inline]{Check that the same terms are used throughout, for example multi-objective vs. multiobjective, optimization vs. optimisation, ...}

\section{Introduction}
As is the case in most empirical research, the optimization community employs benchmarks to compare the performance of algorithms. Unfortunately, issues in both the design and the use of benchmarks are common. These can, for instance, relate to how well benchmarks are able to differentiate between algorithms, but also to how well they reflect properties of real-world application problems. For example, Ishibuchi et al. mention that the performance of an algorithm on popular benchmark problems can be different from that on real-world problems \cite{ishibuchi2019}. Tanabe et al. show that C-DTLZ functions and widely-used real-world-like problems have some unnatural problem features \cite{tanabe2017}. A connection between benchmarks and the properties of real-world problems is important because this allows benchmark results to give a clear idea of how useful different algorithms are in practice. Not only is it valuable to have test functions that imitate real-world problems, some real-world problems may be suitable to be used as a test problem as is, or may have a simplified version that correlates with their true objective(s). The contributor of the problem can benefit from improved solutions and improved algorithms, while academics gain a better understanding of the properties of this optimization problem.

This work focuses on identifying real-world problems and their properties to enable their integration in newly developed benchmark problems. To this end, a questionnaire is developed here to be distributed to specialists working on real-world optimization problems. In addition to learning properties of real-world problems, this work also encourages increased discussion in the optimization community on how to design high quality benchmarks. A few results based on initial responses to the questionnaire are briefly analyzed and discussed. This paper also aims to increase the reach of the questionnaire to get more responses which would allow statistically meaningful conclusions to be drawn.

Besides its obvious relevance to benchmarks, this study is also important for research on optimization algorithms in general. Information obtained through the questionnaire can indicate give an indication about which research directions should be explored more in order to solve problems in the real world. Insights from the questionnaire can thus motivate new directions for research as well as rekindle interest in existing ones, and thus hopefully increase the number of optimization algorithms that are relevant for applications in the real world.

\section{Questionnaire}
\label{sec:questionnaire}
\begin{figure}
\centering
\includegraphics[width = \columnwidth, keepaspectratio]{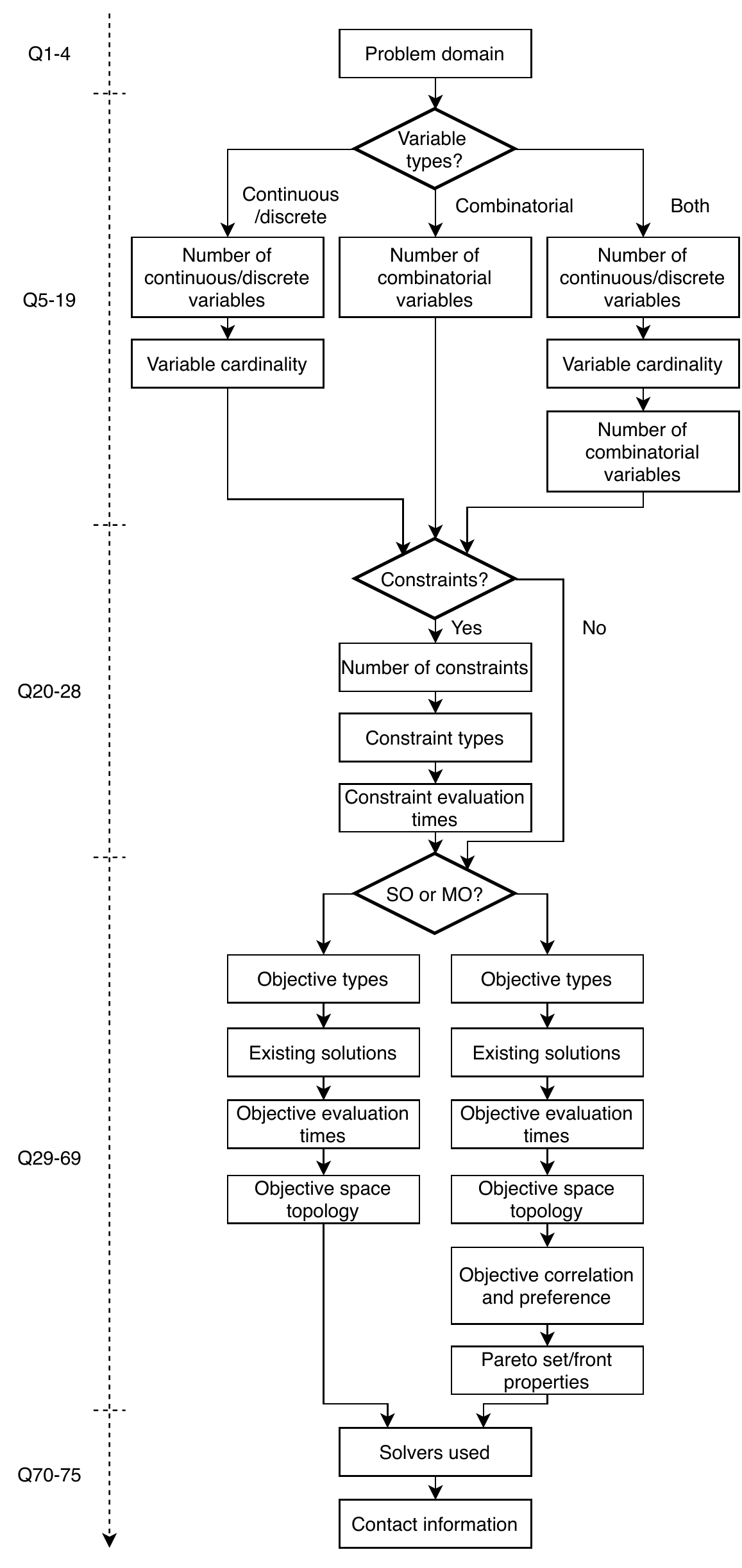}
\caption{Questionnaire structure.}
\label{fig:survey_flow}
\end{figure}

In Figure~\ref{fig:survey_flow} the structure of the questionnaire is given. While there are a total of 75 questions, depending on the route, only between 27 and 53 questions are actually posed.

\section{Discussion and outlook}
\label{sec:conclusion}
Through a newly proposed questionnaire this work aims to improve the understanding of the properties of real-world problems in order to improve the quality of benchmarks in relation to realistic problems.

The initial set of 21 responses already suggests a few things about real-world problems. Firstly, constrained and continuous optimization problems are most common. Secondly, evaluation times are long and can frequently require hours of computation. Thirdly, topological characteristics of a problem's objective space are often unknown. This last point makes designing realistic benchmarks challenging, while the second point requires attention to mitigate evaluation time in a benchmarking setting.

In order to gather more results, and improve the accuracy of any conclusions about the properties of real-world problems, we invite you to participate in the questionnaire at: \url{https://tinyurl.com/opt-survey}.

\begin{acks}
The questionnaire was first proposed at the Lorentz Center MACODA (Many Criteria Optimization and Decision Analysis) workshop as a group effort by 
J.\ Fieldsend, 
J.\ Forde, 
H.\ Ishibuchi, 
E.\ Marescaux, 
M.\ Miyakawa, 
R.\ Purshouse, 
J.\ Richter,
D.\ Thierens,
C.\ Tour{\'e},
and the authors of this paper. We thank the working group participants for their valuable contributions.
We also wish to thank 
M.\ Balvert, 
A.\ Bouter, 
K.\ Chiba,
D.\ Gaudrie, 
M.\ Kanazaki, 
T.\ Kohira,
P.\  Z.\  Korondi,
M.\ van der Meer, 
J.\ Rohmer, 
and all anonymous contributors for their time and effort to fill in the questionnaire.

Timo Deist is funded by the Open Technology Programme (proj. nr. 15586) financed by the Dutch Research Council (NWO), Elekta, Xomnia and co-funded by the public-private partnership allowance for top consortia for knowledge and innovation (TKIs) from the Ministry of Economic Affairs.
Furthermore, Tea Tu{\v s}ar acknowledges financial support from the Slovenian Research Agency (projects No. Z2-8177 and BI-DE/20-21-019). Boris Naujoks acknowledges the 
H2020-MSCA-ITN-2016 UTOPIAE (grant agreement No. 722734) and the DAAD (German Academic Exchange Service), Project-ID: 57515062: "Multi-objective Optimization for Artificial Intelligence Systems in Industry".
\todo[inline]{Add any other (funding) acknowledgments by the authors.}
\end{acks}

\bibliographystyle{ACM-Reference-Format}
\bibliography{sample-base}

\end{document}